\documentclass[10pt,twocolumn,letterpaper]{article}

\usepackage{cvpr}
\usepackage{times}
\usepackage{epsfig}
\usepackage{graphicx}
\usepackage{amsmath}
\usepackage{amssymb}
\usepackage{cite}
\usepackage{subfigure}
\usepackage{color}
\usepackage{footnote}
\makesavenoteenv{table}

\usepackage[pagebackref=true,breaklinks=true,letterpaper=true,colorlinks,bookmarks=false]{hyperref}

\cvprfinalcopy 

\def\cvprPaperID{135} 

\ifcvprfinal\fi
\begin{document}

\title{Learning Deep Context-aware Features over Body and Latent Parts \\ for Person Re-identification}

\author{
Dangwei Li$^{1,2}$, Xiaotang Chen$^{1,2}$, Zhang Zhang$^{1,2}$, Kaiqi Huang$^{1,2,3}$\\
$^{1}$CRIPAC$\;\&\;$NLPR, CASIA  \quad $^{2}$University of Chinese Academy of Sciences\\
$^{3}$CAS Center for Excellence in Brain Science and Intelligence Technology\\
{\tt\small {\{dangwei.li, xtchen, zzhang, kaiqi.huang\}}@nlpr.ia.ac.cn}
}


\maketitle
\thispagestyle{empty}

\begin{abstract}
  Person Re-identification (ReID) is to identify the same person across different cameras.
  It is a challenging task due to the large variations in person pose, occlusion, background clutter, \etc
  How to extract powerful features is a fundamental problem in ReID and is still an open problem today.
  In this paper, we design a Multi-Scale Context-Aware Network (MSCAN) to learn powerful features
  over full body and body parts, which can well capture the local context knowledge by stacking multi-scale convolutions in each layer.
  Moreover, instead of using predefined rigid parts, we propose to learn and localize
  deformable pedestrian parts using Spatial Transformer Networks (STN) with novel spatial constraints.
  The learned body parts can release some difficulties, \eg pose variations and background clutters, in part-based representation.
  Finally, we integrate the representation learning processes of full body and body parts into a unified framework for person ReID
  through multi-class person identification tasks.
  Extensive evaluations on current challenging large-scale person ReID datasets, including the image-based Market1501, CUHK03 and sequence-based MARS datasets,
  show that the proposed method achieves the state-of-the-art results.
\end{abstract}

\vspace{-1.0em}
\section{Introduction}
Person re-identification aims to search for the same person across different cameras with a given probe image.
It has attracted much attention in recent years due to its importance in many practical applications, such as video surveillance and content-based image retrieval.
Despite of years of efforts, it still has many challenges, such as large variations in person pose, illumination, and background clutter.
In addition, similar appearance of clothes among different people and imperfect pedestrian detection results further increase its difficulty in
real applications.

\begin{figure}[tbp]
  \centering
  \includegraphics[width=0.35\paperwidth]{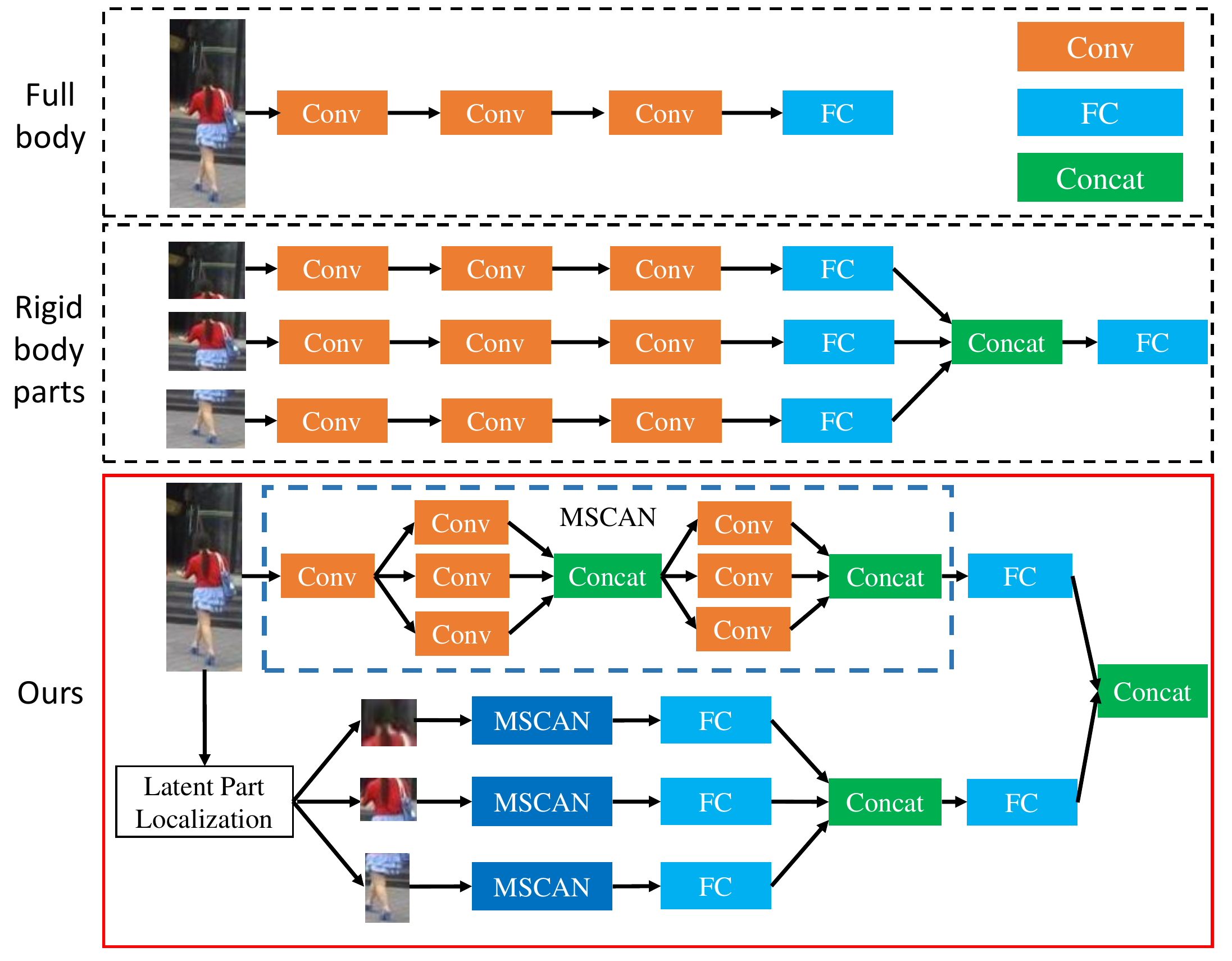}
  \vspace{-0.0em}
  \caption{The schematic of typical feature learning framework with deep learning.
  As shown in black dashed boxes, current approaches focus on the full body or rigid body parts for feature learning.
  Different from them, we use the spatial transformer networks to learn and localize pedestrian parts and use
  multi-scale context-aware convolutional networks to extract full-body and body-parts representations for ReID. Best viewed in color.
  }
  \label{fig:framework_simple}
  \vspace{-1.5em}
\end{figure}

Most existing methods for ReID focus on developing a powerful representation to handle the variations of viewpoint, body pose, background clutter, \etc~\cite{xu2014person,GrayECCV08,FarenzenaCVPR10,YangyangECCV14,Yang2017aaai,KviatkovskyPAMI13color,ZhaoruiCVPR13unsupervised,zhaoruiCVPR14learning,LiaoshengcaiCVPR15,MatsukawaCVPR16,LidangweiArxiv16}, or learning an effective distance metric~\cite{KostingerCVPR12,ProsserBMVC10person,ZhengweishiPAMI13reid,LizhenCVPR13,
LiaoshengcaiCVPR15,ZhangLiCVPR16,ChendapengCVPR16similarity}.
Some of existing methods learn both of them jointly~\cite{LiWeiCVPR14,YiICPR14DML,AhmedCVPR15improved,Shihanlin2016Embedding}.
Recently, deep feature learning based methods~\cite{DingPR15deep,Chengde2016person,Varior2016Siamese,VariorECCV16Gated},
which learn a global pedestrian feature and use Euclidean metric to measure two samples, have obtained the state-of-the-art results.
With the increasing sample size of ReID dataset, the learning of features from multi-class person identification
tasks~\cite{XiaotongCVPR16Domain,ZhengliangECCV16,XiaotongARXIV16end,ZhengliangArxiv16,SchumannArxiv16deep},
denoted as ID-discriminative Embedding (IDE)~\cite{ZhengliangArxiv16},
has shown great potentials on current large-scale person ReID datasets, such as MARS~\cite{ZhengliangECCV16} and PRW~\cite{ZhengliangArxiv16},
where the IDE features are taken from the last hidden layer of Deep Convolutional Neural Networks (DCNN).
In this paper, we aim to learn the IDE feature for person ReID using DCNN.

Existing DCNN models for person ReID typically learn a global full-body representation for input person image (Full body in Figure~\ref{fig:framework_simple}),
or learn a part-based representation for predefined rigid parts (Rigid body parts in Figure~\ref{fig:framework_simple})
or learn a feature embedding for both of them.
Although these DCNN models have obtained impressive results on existing ReID datasets, there are still two problems.
\textbf{First}, for feature learning, current popular DCNN models typically stack single-scale convolution and max pooling layers to generate deep networks.
With the increase of the number of layers, these DCNN models could easily miss some small scale visual cues, such as sunglasses and shoes.
However, these fine-grained attributes are very useful to distinguish the pedestrian pairs with small inter-class variations.
Thus these DCNN models are not the best choice for pedestrian feature learning.
\textbf{Second}, due to the pose variations and imperfect pedestrian detectors, the pedestrian image samples may be misaligned.
Sometimes they may have some backgrounds or lack some parts, \eg legs.
In these cases, for part-based representation, the predefined rigid grids may fail to capture correct correspondence between two pedestrian images.
Thus the rigid predefined grids are far from robust for effective part-based feature learning.

In this paper, we propose to learn the features of full body and body parts jointly.
\textbf{To solve the first problem}, we propose a Multi-Scale Context-Aware Network (MSCAN).
As shown in Figure~\ref{fig:framework_simple}, for each convolutional layer of the MSCAN,
we adopt multiple convolution kernels with different receptive fields to obtain multiple feature maps.
Feature maps from different convolution kernels are concatenated as current layer's output.
To decrease the correlations among different convolution kernels, the dilated convolution~\cite{YuKoltun2016} is used rather than general convolution kernels.
Through this way, multi-scale context knowledge is obtained at the same layer.
Thus the local visual cues for fine-grained discrimination is enhanced.
In addition, through embedding contextual features layer-by-layer (convolution operation across layers),
MSCAN can obtain more context-aware representation for input image.
\textbf{To solve the second problem}, instead of using rigid body parts, we propose to localize latent pedestrian parts through
Spatial Transform Networks (STN)~\cite{JaderbergNIPS15spatial}, which is originally proposed to learn image transformation.
To adapt it to the pedestrian part localization task, we propose three new constraints on the learned transformation parameters.
With these constraints, more flexible parts can be localized at the informative regions, so as to reduce the distraction of background contents.

Generally, the features of full body and body parts are complementary to each other.
The full-body features pay more attention to the global information while the body-part features care more about the local regions.
To better utilize these two types of representations, in this paper, features of full body and body parts are concatenated to form the final pedestrian
representation.
In test stage, the Euclidean metric is adopted to compute the distance between two L2 normalized person representations for person ReID.

The contributions of this paper are summarized as follows:
(a) We propose a multi-scale context-aware network to enhance the visual context information for better feature representation of fine-grained visual cues.
(b) Instead of using rigid parts, we propose to learn and localize pedestrian parts using spatial transformer networks with novel prior spatial constraints.
Experimental results show that fusing the global full-body and local body-part representations greatly improves the performance of person ReID.

\vspace{-0.5em}
\section{Related Work}
\label{relatedwork}
Typical person ReID methods focus on two key points: developing a powerful feature for image representation and
learning an effective metric to make the same person be close and different persons far away.
Recently, deep learning approaches have achieved the state-of-the-art results for person ReID~\cite{XiaotongCVPR16Domain,ZhengliangECCV16,VariorECCV16Gated,zheng2016personreview,zhang2015bit}.
Here we mainly review the related deep learning methods.

Deep learning approaches for person ReID tend to learn person representation and similarity (distance) metric jointly.
Given a pair of person images, previous deep learning approaches learn each person's features followed by a deep matching function
from the convolutional features~\cite{LiWeiCVPR14,AhmedCVPR15improved,Chen2017aaai,Chen2017cvprid} or the Fully Connected (FC) features~\cite{YiICPR14DML,wang2016dari,Shihanlin2016Embedding}.
In addition to the deep metric learning, some work directly learns image representation through pair-wise contrastive loss or triplet ranking loss,
and use Euclidean metric for comparison~\cite{DingPR15deep,Chengde2016person,Varior2016Siamese,VariorECCV16Gated}.

With the increasing sample size of ReID dataset, the IDE feature which is learned through multi-class person identification tasks,
has shown great potentials on current large-scale person ReID datasets.
Xiao \etal~\cite{XiaotongCVPR16Domain} propose the domain guided dropout to learn features over multiple datasets simultaneously with identity classification loss.
Zheng \etal~\cite{ZhengliangECCV16} learn the IDE feature for the video-based person re-identification.
Xiao \etal~\cite{XiaotongARXIV16end} and Zheng \etal~\cite{ZhengliangArxiv16} learn the IDE feature to jointly solve the pedestrian detection and person ReID tasks.
Schumann \etal~\cite{SchumannArxiv16deep} learn the IDE feature for domain adaptive person ReID.
The similar phenomenon has also been validated on face recognition~\cite{SunyiCVPR14deep1}.

As we know, previous DCNN models usually adopt the layer-by-layer single-scale convolution kernels to learn the context information.
Some DCNN models~\cite{YiICPR14DML,Chengde2016person,Shihanlin2016Embedding} adopt rigid body parts to learn local pedestrian features.
Different from them, we improve the classical models in two ways.
Firstly, we propose to enhance the context knowledge through multi-scale convolutions at the same layer.
The relationship among different context knowledge are learned by embedding feature maps layer-by-layer (convolution or FC operation).
Secondly, instead of using rigid parts, we utilize the spatial transformer networks with proposed prior constraints to learn and localize latent human parts.

\begin{figure*}[!htbp]
  \centering
  \includegraphics[width=0.70\paperwidth]{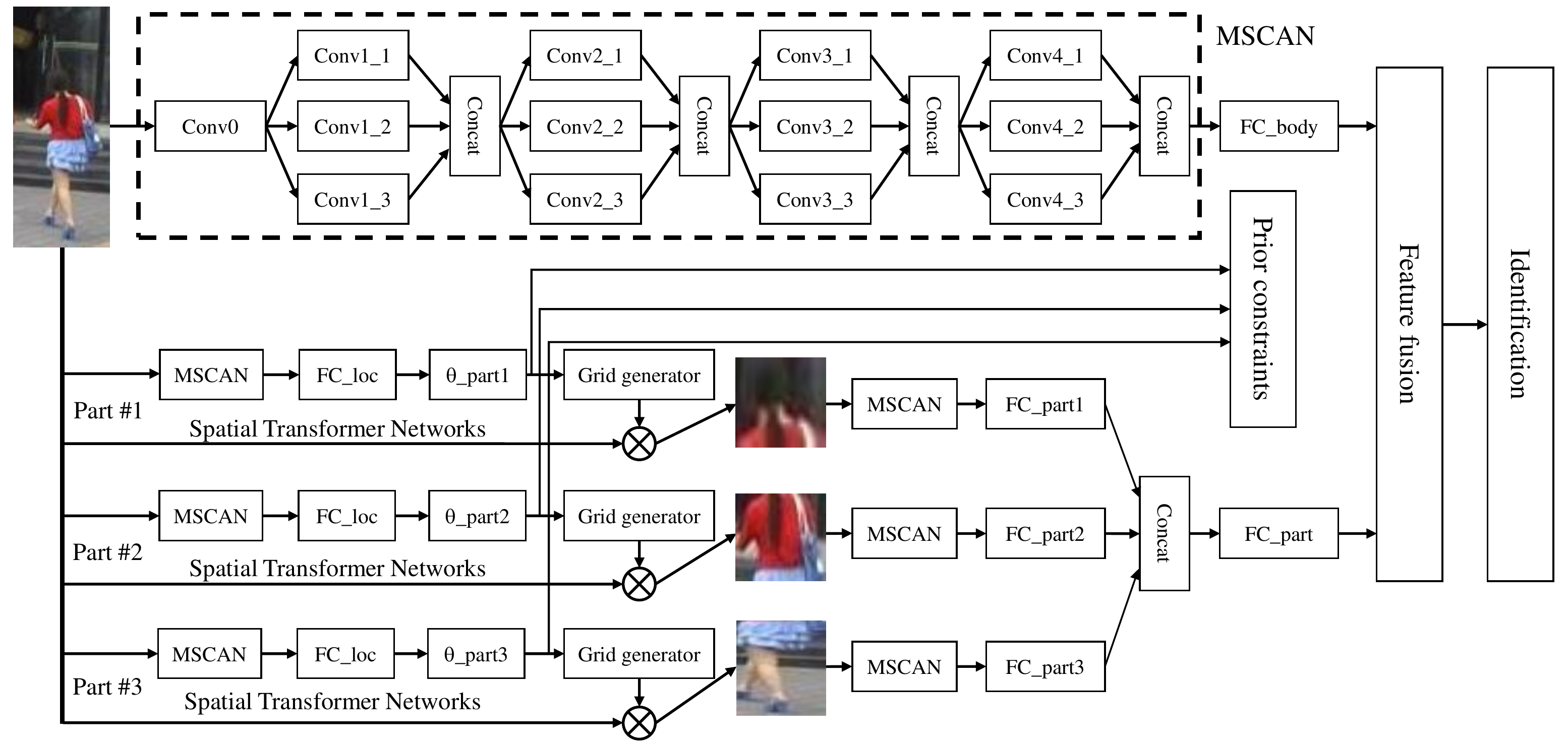}
  \caption{Overall framework of the proposed model. The proposed model consists three components:
  the global body-based feature learning with MSCAN, the latent pedestrian parts localization with spatial
  transformer networks and local part-based feature embedding, the fusion of full body and
  body parts for multi-class person identification tasks.
  }
  \label{fig:framework_all}
  \vspace{-1em}
\end{figure*}

\section{Proposed Method}
\label{proposedmethod}

The focus of this approach is to learn powerful feature representations to describe pedestrians.
The overall framework of the proposed method is shown in Figure~\ref{fig:framework_all}.
In this section, we introduce our model from four aspects: a multi-scale context-aware network for efficient feature learning (Section~\ref{MSCAN}),
the latent parts learning and localization for better local part-based feature representation (Section~\ref{LatentPartLoc}),
the fusion of global full-body and local body-part features for person ReID (Section~\ref{FeatureFusion}),
and our final objective function in Section~\ref{ObjectiveFunction}.

\subsection{Multi-scale Context-aware Network}
\label{MSCAN}
Visual context is an important component to assist visual-related tasks, such as object recognition~\cite{LintsungyiECCV14microsoft} and object detection~\cite{ZhengPAMI12context,ZengArxiv16crafting}.
Typical convolutional neural networks model context information through hierarchical convolution and pooling~\cite{KrizhevskyNIPS12,HekaimingCVPR16Residual}.
For person ReID task, the most important visual cues are visual attribute knowledge, such as clothes color and types.
However, they have large variations in scale, shape and position, such as the hat/glasses at small local scale and the cloth color at the larger scale.
Directly using bottom-to-up single-scale convolution and pooling may not be effective to handle these complex variations.
Especially, with the increase number of layers, the small visual regions, such as hat, will be easily missed in top layers.
To better learn these diverse visual cues, we propose the Multi-scale Context-Aware Network.

\begin{table}[!hbtp]
  \begin{center}
  \scriptsize
    \begin{tabular}{|l|c|c|c|c|c|}
    \hline
    layer & dilation & kernel & pad   & \#filters & output \\
    \hline
    input &    -   &   -    &  -     &   -    & 3$\times$160$\times$64 \\
    \hline
    conv0 & 1     & 5$\times$5   & 2     & 32    & 32$\times$160$\times$64 \\
    \hline
    pool0 & -     & 2$\times$2    & -     & -     & 32$\times$80$\times$32 \\
    \hline
    conv1 & 1/2/3 & 3$\times$3   & 1/2/3 & 32/32/32 & 96$\times$80$\times$32 \\
    \hline
    pool1 & -     & 2$\times$2     & -     &    -   & 96$\times$40$\times$16 \\
    \hline
    conv2 & 1/2/3 & 3$\times$3   & 1/2/3 & 32/32/32 & 96$\times$40$\times$16 \\
    \hline
    pool2 & -     & 2$\times$2     & -     &    -   & 96$\times$20$\times$8 \\
    \hline
    conv3 & 1/2/3 & 3$\times$3   & 1/2/3 & 32/32/32 & 96$\times$20$\times$8 \\
    \hline
    pool3 & -     & 2$\times$2     & -     &    -   & 96$\times$10$\times$4 \\
    \hline
    conv4 & 1/2/3 & 3$\times$3   & 1/2/3 & 32/32/32 & 96$\times$10$\times$4 \\
    \hline
    pool4 & -     & 2$\times$2     & -     &    -   & 96$\times$5$\times$2 \\
    \hline
    \end{tabular}%
  \end{center}
  \caption{Model architecture of MSCAN.}
  \label{tab:mscan}%
  \vspace{-1em}
\end{table}%

The architecture of the proposed MSCAN is shown in Tabel~\ref{tab:mscan}.
It has an initial convolution layer with kernel size $5\times5$ to capture the low-level visual features.
Then we use four multi-scale convolution layers to obtain the complex image context information.
In each multi-scale convolution layer, we use a convolution kernel with size $3\times3$.
To obtain multi-scale receptive fields, we adopt dilated convolution~\cite{YuKoltun2016} for the convolution filters.
We use three different dilation ratios, i.e. 1,2 and 3, to capture different scale context information.
The feature maps from different dilation ratios are concatenated along the channel axis to form the final output of the current convolution layer.
Thus, the visual context information are enhanced explicitly.
To integrate different context information together, the feature maps of current convolution layer are embedded through layer-by-layer convolution or FC operation.
As a result, the visual cues at different scales are fused in a latent way.
Besides, we adopt Batch Normalization~\cite{Ioffe15batch} and ReLU neural activation units after each convolution layer.

\begin{figure}[!tbp]
  \centering
  \includegraphics[width=0.35\paperwidth]{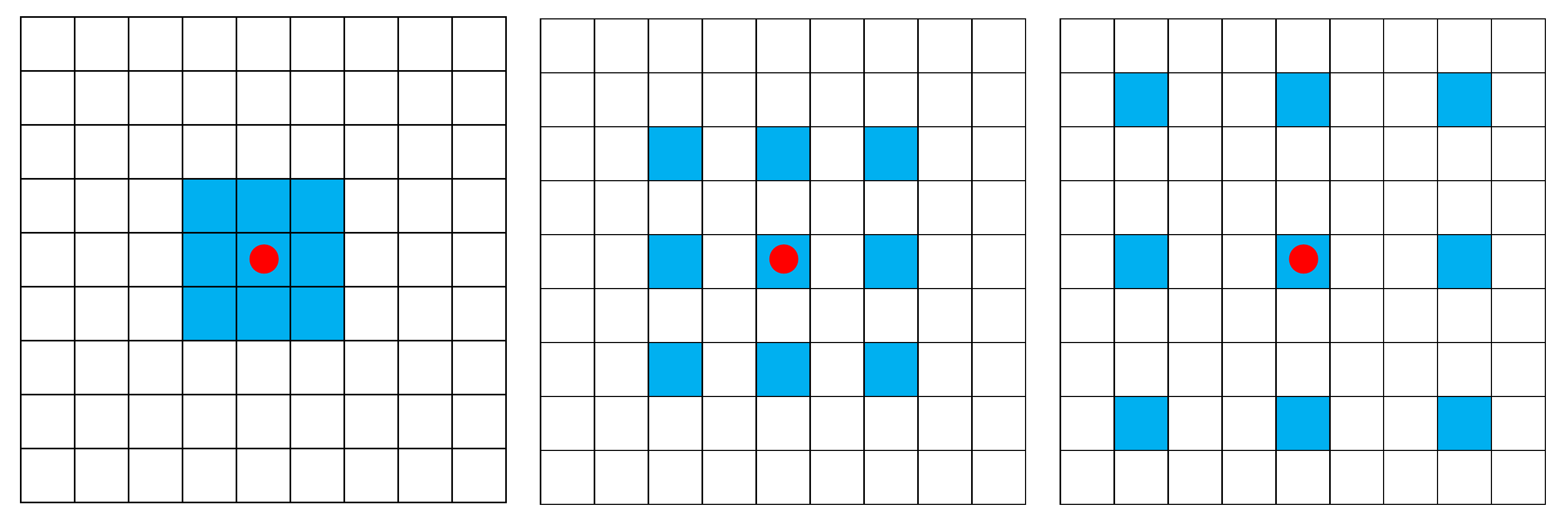}
  \caption{Example of dilated convolution for the same input feature map.
  The convolutional kernel is $3\times3$ and the dilation ratio from left to right is 1, 2, and 3.
  The blue boxes are effective positions for convolution at the red circle.
  Best viewed in color.
  }
  \label{fig:DilationCov}
  \vspace{-1em}
\end{figure}

In this paper, we use the dilated convolutions with dilation ratios 1, 2 and 3 instead of the classic convolution filters with kernel
size $3\times3$, $5\times5$ and $7\times7$.
The main reason is that the classic convolution filters with kernel size $3\times3$, $5\times5$ and $7\times7$ overlap with each other at the same output position and produce redundant information.
To make it clearer, we show the dilated convolution kernel (size $3\times3$) with dilation ratio ranging from $1$ to $3$ in Figure~\ref{fig:DilationCov}.
For the same output position which is shown in red circle, the convolution kernel with larger dilation ratio has larger receptive field, while only the center position is overlapped with other convolution kernels.
This can reduce the redundant information among filters with different receptive fields.

In summary, as shown in Figure~\ref{fig:framework_all}, we use MSCAN to learn the multi-scale context representation for full body and body parts.
In addition, it is also used for feature learning in spatial transformer networks mentioned below.

\subsection{Latent Part Localization}
\label{LatentPartLoc}
Pedestrian parts are important in person ReID.
Some existing work~\cite{GrayECCV08,LiaoshengcaiCVPR15,YiICPR14DML,Chengde2016person} has explored rigid body parts to develop robust features.
However, due to the unsatisfying pedestrian detection algorithms and large pose variations, the method of using rigid body parts for local
feature learning is not the optimal solution.
As shown in Figure~\ref{fig:framework_simple}, when using rigid body parts, the top part consists of large amount of background.
This motivates us to learn and localize the pedestrian parts automatically.

We integrate STN~\cite{JaderbergNIPS15spatial} as the part localization net in our proposed model.
The original STN is proposed to explicitly learn the image transformation parameters, such as translation and scale.
It has two main advantages: (1) it is fully differentiable and can be easily integrated into existing deep learning frameworks,
(2) it can learn to translate, scale, crop or warp an interesting region without explicit region annotations.
These facts make it very suitable for pedestrian parts localization.

STN includes two components, the spatial localization network to learn the transformation parameters,
and the grid generator to sample the input image using an image interpolation kernel. 
More details about STN can be seen in \cite{JaderbergNIPS15spatial}.
In our implementation of STN, the bilinear interpolation kernel is adopted to sample the input image.
And four transformation parameters $\theta=[s_x, t_x, s_y, t_y]$ are used, where $s_x$ and $s_y$ are the horizontal and vertical scale transformation parameters,
and $t_x$ and $t_y$ are the horizontal and vertical translation parameters.
The image height and width are normalized to be in $[-1, 1]$.
Only scale and translation parameters are learned because these two types of transformations serve enough to crop the pedestrian parts effectively.
The transformation is applied as an inverse warping to generate the output body part regions:
\vspace{-0.2em}
\begin{gather}
\begin{pmatrix} x^{in}_{i} \\ y^{in}_{i} \end{pmatrix} =
\begin{bmatrix} s_x & 0 & t_x \\ 0 & s_y & t_y \end{bmatrix}
\begin{pmatrix} x^{out}_{i} \\ y^{out}_{i} \\ 1 \end{pmatrix}
\end{gather}

where $x^{in}$ and $y^{in}$ are the input image coordinates, $x^{out}$ and $y^{out}$ are the output part image coordinates, and $i$ indexes the pixels in the output body part image.

In this paper, we expect STN to learn three parts corresponding to the head-shoulder, upper body and lower body.
Each part is learned by an independent STN from the original pedestrian image.
For the spatial localization network, firstly we use MSCAN to extract the global image feature maps.
Then we learn the high-level abstract representation by a 128-dimension FC layer (FC\_{loc} in Figure~\ref{fig:framework_all}).
At last, we learn the transformation parameters $\theta$ with a 4-dimension FC layer based on the FC\_loc.
The MSCAN and FC\_{loc} are shared among three spatial localization networks.
The grid generator can crop the learned pedestrian parts based on the learned transformation parameters.
In this paper, the resolution of the cropped part image is $96\times64$.

For the part localization network, it is hard to learn three groups of parameters for part localization.
There are three problems.
First, the predicted parts from STN can easily fall into the same region, \eg, the center region of a person,
and result in redundance.
Second, the scale parameters can easily become negative and the pedestrian part will be mirrored vertically or horizontally or both.
This is not consistent with general human cognition.
Because few person will stand upside down in surveillance scenes.
At last, the cropped parts may fall out of the person image, thus the network would be hard to converge.
To solve the above problems, we propose three prior constraints on the transformation parameters in the part localization network.

The first constraint is for the positions of predicted parts.
We expect the predicted parts to be near the prior center points, so that the learned parts would be complementary to each other.
This is termed as the center constraint, which is formalized as follows:
\begin{equation}
L_{cen} = \frac{1}{2}\max\{0, (t_x-C_x)^2 + (t_y-C_y)^2 - \alpha\}
\end{equation}
where $C_x$ and $C_y$ are prior center points for each part.
$\alpha$ is the threshold to control the translation between estimated and prior center points.
In our experiments, we set the prior center point ($C_x, C_y$) to $(0, 0.6)$, $(0, 0)$, and $(0, -0.6)$ for each part. The threshold $\alpha$ is set to $0.5$.

The second one is the value range constraint on the predicted scale parameter.
We hope the scale to be positive, so that the predicted parts have a reasonable extent.
The value range constraint on the scale parameter is formalized as follows:
\begin{equation}
L_{pos} = \max\{0, \beta - s_x \} + \max\{0, \beta - s_y \}
\end{equation}
where $\beta$ is threshold parameter and is set to 0.1 in this paper.

The last one is to make the localization network focus on the inner region of an image.
It is formalized as follows:
\begin{equation}
\renewcommand\arraystretch{1.5}
\begin{array}{r}
L_{in} = \frac{1}{2}\max\{0, ||s_x \pm t_x||^2 - \gamma \} \\
+ \frac{1}{2}\max\{0, ||s_y \pm t_y||^2 - \gamma \}
\end{array}
\end{equation}
where $\gamma$ is the boundary parameter. $\gamma$ is set to 1.0 in our paper, which means the cropped parts should be inside the pedestrian image.

Finally the loss for the transformation parameters in the part localization network is described as follows:
\begin{equation}
\label{equ:los_loc}
L_{loc} = L_{cen} + \xi_1 L_{pos} + \xi_2 L_{in}
\end{equation}
where $\xi_1$ and $\xi_2$ are hyperparameters.
The hyperparameters $\xi_1$ and $\xi_2$ are both set to 1.0 in our experiments.

\subsection{Feature Extraction and Fusion}
\label{FeatureFusion}
The features of full body and body parts are learned by separate networks and then are fused in a unified framework for multi-class person identification tasks.
For the body-based representation, we use MSCAN to extract the global feature maps and then learn a 128-dimension feature embedding (denoted as FC\_body in Figure~\ref{fig:framework_all}).
For the part-based representation, first, for each body part, we use the MSCAN to extract its feature maps and learn a 64-dimension feature embedding (denoted as FC\_part1, FC\_part2, FC\_part3).
Then, we learn a 128-dimension feature embedding (denoted as FC\_part) based on features of each body part.
The Dropout~\cite{srivastava2014dropout} is adopted after each FC layer to prevent overfitting.
At last, the features of global full body and local body parts are concatenated to be a 256-dimension feature as the final person representation.

\subsection{Objective Function}
\label{ObjectiveFunction}

In this paper, we adopt the softmax loss as the objective function for multi-class person identification tasks.
\begin{equation}
\label{equ:los_cls}
L_{cls} = -\sum_{i=1}^{N}log\frac{\exp(W_{y_i}^{T}x_i+b_{y_i})}{\sum\nolimits_{j=1}^{C}\exp(W_j^Tx_i+b_j)}
\end{equation}
where $i$ is the index of person images, $x_i$ is the feature of $i$-th sample, $y_i$ is the identity of $i$-th sample,
$N$ is the number of person images, $C$ is the number of person identities, $W_j$ is the classifier for $j$-th identity.

For the overall network training, we use the classification and localization loss jointly.
The final objective function is as follows.
\begin{equation}
\label{equ:los_all}
L = L_{cls} + \lambda L_{loc}
\end{equation}
where the $\lambda$ is the hyperparameter, which is set to 0.1 in our experiments.

\section{Experiments}
\label{experiments}
In this paragraph, the datasets and evaluation protocols are introduced in Section~\ref{exp:dataset}.
Implementation details are described in Section~\ref{exp:detail}.
Comparisons with state-of-the-art methods are discussed in Section~\ref{exp:stateoftheart}.
The effectiveness of proposed model is analyzed in Section~\ref{exp:effmscan} and Section~\ref{exp:efflpl}.
Cross-dataset evaluation is described in Section~\ref{exp:discussion}.

\subsection{Datasets and Protocols}
\label{exp:dataset}
\textbf{Datasets.} In this paper, we evaluate our proposed method on current largest person ReID datasets,
including Market1501~\cite{ZhengliangICCV15}, CUHK03~\cite{LiWeiCVPR14} and MARS~\cite{ZhengliangECCV16}.
We do not directly train our model on small datasets, such as VIPeR~\cite{Gray07VIPeR}.
It would be easily overfitting and insufficient to learn such a large capacity network on small datasets from scratch.
However, we give some results through fine-tuneing the model from Market1501 to VIPeR and make cross-dataset ReID on VIPeR for generalization validation.
Related experimental results are discussed in Section~\ref{exp:discussion}.


Market1501: It contains 1,501 identities which are captured by six manually set cameras.
There are 32,368 pedestrian images in total.
Each person has 3.6 images on average at each viewpoint.
It provides two types of images, including cropped and automatically detected pedestrians by the Deformable Part based Model (DPM)~\cite{FelzenszwalbPAMI10object}.
Following ~\cite{ZhengliangICCV15}, 751 identities are used for training and the rest 750 identities are used for testing.

CUHK03: It contains 1,360 identities which are captured by six surveillance cameras in campus.
Each identity is captured by two disjoint cameras.
Totally it consists of 13,164 person images and each identity has about 4.8 images at each viewpoint.
This dataset provides two types of annotations, including manually annotated bounding boxes, and bounding boxes detected using DPM.
We validate our proposed model on both types of data.
Following~\cite{LiWeiCVPR14}, we use 1,260 person identities for training and the rest 100 identities for testing.
Experiments are conducted 20 times and the mean result is reported.

MARS: It is the largest sequence-based person ReID dataset.
It contains 1,261 identities with each identity captured by at least two cameras.
It consists of 20,478 tracklets and 1,191,003 bounding boxes.
Following~\cite{ZhengliangECCV16}, we use 625 identities for training and the rest 631 identities for testing.

\textbf{Protocols.}
Following original evaluation protocols in each dataset, we adopt three evaluation protocols for fair comparison with existing methods.
The first one is Cumulated Matching Characteristics (CMC) which is adopted on the CUHK03 and MARS datasets.
The second one is Rank-1 identification rate on the Market1501 dataset.
The third one is mean Average Precision (mAP) on the Market1501 and MARS datasets.
mAP considers both precision and recall rate, which could be complementary to CMC.

\subsection{Implementation Details}
\label{exp:detail}
\textbf{Model:} We try to learn the pedestrian representation through multi-class person identification tasks using full body and body parts.
To evaluate the effectiveness of full body and body parts independently, we extract two sub-models from the whole network of Figure~\ref{fig:framework_all}.
The first one only uses the full body to learn the person representation with identity classification loss.
The second one only uses the parts to learn the person representation with identity classification and body parts localization loss.
For person re-identification, we use the L2 normalized person representation and Euclidean metric to measure the distance between two pedestrian samples.

\textbf{Optimization:}
Our model is implemented based on Caffe~\cite{JiaMM14caffe}.
We use all the available training identities for training and randomly select one sample for each identity for validation.
As the dataset can be quite large, in practice we use a stochastic approximation of the objective function.
Training data is randomly divided into mini-batches with a batch size of 64.
The model performs forward propagation on each mini-batch and computes the loss.
Backpropagation is then used to compute the gradients on each mini-batch and the weights are updated with stochastic gradient descent.
We start with a base learning rate of $\eta = 0.01 $ and gradually decrease it after each $1\times10^4$ iterations.
It should be noted that the learning rate of part localization network is 1\% of that in feature learning network.
We use a momentum of $\mu = 0.9$ and weight decay $\lambda = 5\times10^{-3}$.
For overall network training, we initialize the network using pretrained body-based and part-based model and then follow the same training strategy as described above.
We use the model at $5\times10^4$ iterations for testing.

\textbf{Data Preprocessing:} For each image, we resize it to $160\times64$, subtract the mean value on each channel (B, G and R), and then normalize it with scale $1.0/256$ for network training.
To prevent overfitting, we randomly reflect each image horizontally in the training stage.

\subsection{Comparison with State-of-the-art Methods}
\label{exp:stateoftheart}

\textbf{Market1501:} For the Market1501 dataset, several state-of-the-art methods are compared, including Bag of Words (BOW)~\cite{ZhengliangICCV15}, Weighted Approximate Rank Component Analysis (WARCA)~\cite{Jose2016scalable}, Discriminative Null Space (DNS)~\cite{ZhangLiCVPR16}, Spatially Constrained Similarity function on Polynomial feature map (SCSP)~\cite{ChendapengCVPR16similarity}, and deep learning based approaches, such as PersonNet~\cite{Wulin2016Personnet}, Comparative Attention Network (CAN)~\cite{Liu2016end}, Siamese Long Short-Term Memory (S-LSTM)~\cite{Varior2016Siamese}, Gated Siamese Convolutional Neural Network (Gate-SCNN)~\cite{VariorECCV16Gated}.
The experimental results are shown in Table~\ref{tab:marketresults}.

Compared with existing full body-based convolutional neural networks, such as CAN and Gate-SCNN, the proposed network structure can better capture pedestrian features with multi-class person identification tasks.
Our full-body representation improves Rank-1 identification rate by 9.57\% on the state-of-the-art results produced by the Gate-CNN in single query.
Compared with the full body, our body-part representation increase 0.80\%.
The main reason is that the pedestrians detected by DPM consists much more background information and the part-based representation can better reduce
the influences of background clutter.

The full-body and body-part representations are complementary to each other.
The full-body representation cares more about the global information, such as the background and body shape.
The body-part representation pays more attention to parts, such as head, upper body and lower body.
As shown in Table~\ref{tab:marketresults}, the fusion model of full body and body parts improves Rank-1 identification rate by more than 4.00\%
compared with the body and parts-based models separately in single query.
The mAP improves about 17.98\% compared with the best result produced by Gate-CNN.

\begin{table}[!tbp]
  \begin{center}
  \scriptsize
    \begin{tabular}{|l|cc|cc|}
    \hline
    Query & \multicolumn{2}{c|}{Single query} & \multicolumn{2}{c|}{Multiple query} \\
    \hline
    Evaluation metrics & R1    & mAP   & R1    & mAP \\
    \hline
    \hline
    BOW~\cite{ZhengliangICCV15}  & 34.38 & 14.1  & 42.64 & 19.47 \\
    BOW + HS~\cite{ZhengliangICCV15} & -     & -     & 47.25 & 21.88 \\
    \hline
    WARCA~\cite{Jose2016scalable} & 45.16 & -     & -     & - \\
    PersonNet~\cite{Wulin2016Personnet} & 37.21 & 26.35 & -     & - \\
    S-LSTM~\cite{Varior2016Siamese} & -     & -     & 61.6  & 35.3 \\
    SCSP~\cite{Chengde2016person}  & 51.9  & 26.35 & -     & - \\
    CAN~\cite{liu2017end}   & 48.24 & 24.43 & -     & - \\
    DNS~\cite{ZhangLiCVPR16}   & 55.43  & 29.87  & 71.56  & 46.03  \\
    Gate-SCNN~\cite{VariorECCV16Gated} & 65.88  & 39.55  & 76.04  & 48.45  \\
    \hline
    \hline
    Our-Part & 76.25  & 53.33  & 84.12  & 62.90    \\
    Our-Body & 75.45  & 52.41  & 83.43  & 62.03    \\
    Our-Fusion & \textbf{80.31}  & \textbf{57.53}  & \textbf{86.79}  & \textbf{66.70}  \\
    \hline
    \end{tabular}%
  \end{center}
  \vspace{-0.5em}
  \caption{Experimental results on the Market1501 dataset. - means that no reported results are available.}
  \label{tab:marketresults}%
  \vspace{-2em}
\end{table}%

\textbf{CUHK03:} For the CUHK03 dataset, we compare our method with many existing approaches, including Filter Pair Neural Networks (FPNN)~\cite{LiWeiCVPR14}, Improved Deep Learning Architecture (IDLA)~\cite{AhmedCVPR15improved}, Cross-view Quadratic Discriminant Analysis (XQDA)~\cite{LiaoshengcaiCVPR15}, PSD constrained asymmetric metric learning (denoted as MLAPG)~\cite{LiaoshengcaiICCV15},
Sample-Specific SVM (SS)~\cite{ZhangCVPR16sample}, Single image and Cross image representation (SI-CI)~\cite{WangfaqiangCVPR16JSC},
Embedding Deep Metric (EDM)~\cite{Shihanlin2016Embedding}, Domain Guided Dropout (DGD)~\cite{XiaotongCVPR16Domain}, DNS, S-LSTM and Gate-SCNN.
On this dataset, we conduct experiments on both the detected and the labeled datasets.
As presented in previous work~\cite{LiWeiCVPR14}, we use the CMC curve in the single shot case to evaluate the performance.
The overall results are shown in the Table~\ref{tab:cuhk03detectedresults} and Table~\ref{tab:cuhk03labeledresults}.
The full CMC curves are shown in supplementary materials.

Compared with metric learning methods, such as the state-of-the-art approach DNS, the proposed fusion model improves the Rank-1 identification rate by 11.66\% and 13.29\% on the labeled and detected datasets respectively.
Compared with the similar multi-class person identification network DGD, the Rank-1 identification rate improves by 1.63\%
using our fusion model on the labeled dataset.
It should be noted that we only use the labeled sets for training, while the DGD is trained on both the labeled and detected datasets.
This demonstrates the effectiveness of the proposed model.

\begin{table}[!tbp]
  \begin{center}
  \scriptsize
    \begin{tabular}{|l|cccc|}
    \hline
    Dataset & \multicolumn{4}{c|}{CUHK03 detected} \\
    \hline
    Rank  & 1     & 5     & 10    & 20 \\
    \hline
    \hline
    FPNN~\cite{LiWeiCVPR14}  & 19.89  & 50.00  & 64.00  & 78.50  \\
    IDLA~\cite{AhmedCVPR15improved}  & 44.96  & 76.01  & 83.47  & 93.15  \\
    XQDA~\cite{LiaoshengcaiCVPR15}  & 46.25  & 78.90  & 88.55  & 94.25  \\
    MLAPG~\cite{LiaoshengcaiICCV15} & 51.15  & 83.55  & 92.05  & 96.90  \\
    SS-SVM~\cite{ZhangCVPR16sample} & 51.20  & 80.80  & 89.60  & 95.50  \\
    SI-CI~\cite{WangfaqiangCVPR16JSC} & 52.17  & 84.30  & 92.30  & 95.00  \\
    DNS~\cite{ZhangLiCVPR16}   & 54.70  & 84.75  & 94.80  & 95.20  \\
    S-LSTM~\cite{Varior2016Siamese} & 57.30  & 80.10  & 88.30  & - \\
    Gate-SCNN~\cite{VariorECCV16Gated} & 61.80  & 80.90  & 88.30  & - \\
    EDM~\cite{Shihanlin2016Embedding}   & 52.09  & 82.87  & 91.78  & 97.17  \\
    \hline
    \hline
    Our-Part & 62.74  & 88.53  & 93.97  & 97.21  \\
    Our-Body & 64.95  & 89.82  & 94.58  & 97.56  \\
    Our-Fusion & \textbf{67.99}  & \textbf{91.04}  & \textbf{95.36}  & \textbf{97.83}    \\
    \hline
    \end{tabular}%
  \end{center}
  \vspace{-0.5em}
  \caption{Experimental results on the CUHK03 detected dataset.}
  \label{tab:cuhk03detectedresults}%
  \vspace{-0em}
\end{table}%
\begin{table}[!tbp]
  \begin{center}
  \scriptsize
    \begin{tabular}{|l|cccc|}
    \hline
    Dataset & \multicolumn{4}{c|}{CUHK03 labeled} \\
    \hline
    Rank  & 1     & 5     & 10    & 20 \\
    \hline
    \hline
    FPNN~\cite{LiWeiCVPR14}  & 20.65  & 51.50  & 66.50  & 80.00  \\
    IDLA~\cite{AhmedCVPR15improved}  & 54.74  & 86.50  & 93.88  & 98.10  \\
    XQDA~\cite{LiaoshengcaiCVPR15}  & 52.20  & 82.23  & 92.14  & 96.25  \\
    MLAPG~\cite{LiaoshengcaiICCV15} & 57.96  & 87.09  & 94.74  & 98.00  \\
    Ensemble~\cite{PaisitkriangkraiCVPR15} & 62.10  & 89.10  & 94.80  & 98.10  \\
    SS-SVM~\cite{ZhangCVPR16sample} & 57.00  & 85.70  & 94.30  & 97.80  \\
    DNS~\cite{ZhangLiCVPR16}   & 62.55  & 90.05  & 94.80  & 98.10  \\
    EDM~\cite{Shihanlin2016Embedding}   & 61.32  & 88.90  & 96.44  & \textbf{99.94}  \\
    DGD~\cite{XiaotongCVPR16Domain}   & 72.58  & 91.59  & 95.21  & 97.72 \\
    \hline
    \hline
    Our-Part & 69.41  & 92.68  & 96.68  & 99.02  \\
    Our-Body & 71.88  & 93.66  & 97.46  & 99.18  \\
    Our-Fusion & \textbf{74.21}  & \textbf{94.33}  & \textbf{97.54}  & 99.25    \\
    \hline
    \end{tabular}%
  \end{center}
  \vspace{-0.5em}
  \caption{Experimental results on the CUHK03 labeled dataset.}
  \label{tab:cuhk03labeledresults}%
  \vspace{-1em}
\end{table}%

\textbf{MARS:} This dataset is the largest sequence-based person ReID dataset.
On this dataset, we compare the proposed method with several classical methods, including Keep It as Simple and straightforward Metric (KISSME)~\cite{KostingerCVPR12}, XQDA~\cite{LiaoshengcaiCVPR15}, and CaffeNet~\cite{KrizhevskyNIPS12}.
Similar to previous work~\cite{ZhengliangECCV16}, both single query and multiple query are evaluated on MARS.
The overall experimental results on the MARS are shown in Table~\ref{tab:marsresults_single} and Table~\ref{tab:marsresults_mutiple}.
The full CMC curves are shown in supplementary materials.

Compared with CaffeNet, a similar multi-class person identification network, our body-based model
improves the Rank-1 identification rate by 2.93\% and mAP by 4.22\% using XQDA in single query.
It should be noticed that our network does not use any pre-training with additional data.
Usually deep learning network can obtain better results when pretrained with on ImageNet classification task.
Our fusion model improves Rank-1 identification rate and mAP by 6.47\% and by 8.45\% in single query.
This illustrates the effectiveness of our model.
\begin{table}[!tbp]
  \begin{center}
  \scriptsize
    \begin{tabular}{|l|cccc|}
    \hline
    Query & \multicolumn{4}{c|}{Single query} \\
    \hline
    Evaluation metrics & 1     & 5     & 20    & mAP \\
    \hline
    \hline
    CNN+Eulidean~\cite{ZhengliangECCV16} & 58.70  & 77.10  & 86.80  & 40.40  \\
    CNN+KISSME~\cite{ZhengliangECCV16} & 65.00  & 81.10  & 88.90  & 45.60  \\
    CNN+XQDA~\cite{ZhengliangECCV16} & 65.30  & 82.00  & 89.00  & 47.60  \\
    \hline
    \hline
    Our-Fusion+Eulidean & 68.38  & 84.19  & 91.52  & 51.13  \\
    Our-Fusion+KISSME & 69.24  & 85.15  & 92.17  & 53.00  \\
    \hline
    \hline
    Our-Part+XQDA & 66.62  & 82.07  & 90.76  & 49.74  \\
    Our-Body+XQDA & 68.23  & 83.99  & 92.17  & 51.82  \\
    Our-Fusion+XQDA & \textbf{71.77}  & \textbf{86.57}  & \textbf{93.08}  & \textbf{56.05}    \\
    \hline
    \end{tabular}%
    \end{center}
  \caption{Experimental results on the MARS with single query.}
  \vspace{-1em}
  \label{tab:marsresults_single}%
\end{table}%

\begin{table}[!tbp]
  \begin{center}
  \scriptsize
    \begin{tabular}{|l|cccc|}
    \hline
    Query & \multicolumn{4}{c|}{Multiple query} \\
    \hline
    Evaluation metrics & 1     & 5     & 20    & mAP \\
    \hline
    \hline
    CNN+KISSME+MQ~\cite{ZhengliangECCV16} & 68.30  & 82.60  & 89.40  & 49.30  \\
    \hline
    Our-Fusion+Euclidean+MQ & 78.28  & 91.97  & 96.87  & 61.62    \\
    Our-Fusion+KISSME+MQ & 80.51  & 93.18  & 97.22  & 63.50    \\
    Our-Fusion+XQDA+MQ & \textbf{83.03}  & \textbf{93.69}  & \textbf{97.63}  & \textbf{66.43}  \\
    \hline
    \end{tabular}%
    \end{center}
    \vspace{-0.5em}
  \caption{Experimental results on the MARS with multiple query.}
  \vspace{-1em}
  \label{tab:marsresults_mutiple}%
\end{table}%


\subsection{Effectiveness of MSCAN}
\label{exp:effmscan}
To determine the effectiveness of MSCAN, we explore four variants of MSCANs to learn IDE feature based on the whole body image, which is denoted as MSCAN-$k$, $k = \{1,2,3,4\}$.
$k$ is the number of dilation ratios.
For example, MSCAN-$3$ means for each convolution layer in Conv1-Conv4, there are three convolution kernels with dilation ratio 1, 2, and 3 respectively.
With the increase of $k$, the MSCAN captures larger context information at the same convolution layer.

The experimental results based on these four types of MSCANs on the Market1501 dataset are shown in Table~\ref{tab:mscanmarket}.
As we can see, with the increase of the number of dilation ratios, the Rank-1 identification rate and mAP improve stably in single query case.
For multiple query case, which means fusing all images belonging to the same query person at the same camera through average pooling in feature space, the Rank-1 identification rate and mAP also improves step by step.
However, the Rank-1 identification rate and mAP increase not much when $K$ increase from 3 to 4.
We think there is a suitable number of dilation ratios for feature learning.
Considering the model complexity and accuracy improvements in Rank-1 identification rate, we adopt the MSCAN-3 as our final MSCAN model in this paper.
\begin{table}[htbp]
  \begin{center}
    \scriptsize
    \begin{tabular}{|l|cc|cc|}
    \hline
    Query type & \multicolumn{2}{c|}{Single query} & \multicolumn{2}{c|}{Multiple query} \\
    \hline
    Evaluation metrics & Rank-1    & mAP   & Rank-1    & mAP \\
    \hline
    \hline
    MSCAN-1 & 65.38  & 41.85  & 75.21  & 51.14  \\
    MSCAN-2 & 72.21  & 49.19  & 82.22  & 59.03  \\
    MSCAN-3 & 75.45  & 52.41  & 83.43  & 62.03  \\
    MSCAN-4 & \textbf{76.25}  & \textbf{53.14}  & \textbf{84.09}  & \textbf{62.95}   \\
    \hline
    \end{tabular}%
  \end{center}
  \vspace{-0.5em}
  \caption{Experimental results of four types of MSCAN using body-based representation for ReID on the Market1501 dataset.}
  \vspace{-1em}
  \label{tab:mscanmarket}%
\end{table}%

\subsection{Effectiveness of Latent Part Localization}
\label{exp:efflpl}

\textbf{Learned parts \vs rigid parts}
To compare with popular rigid pedestrian parts, we divide the pedestrian into three overlapped regions as predefined rigid parts.
We use the rigid body parts instead of the learned latent body parts for part-based feature learning.
Experimental results with rigid and learned body parts are shown in Table~\ref{tab:partmarket}.
Compared with rigid body parts, the learned body parts improve Rank-1 identification rate and mAP
by 3.27\% and 3.73\% in single query, and by 1.70\% and by 2.67\% in multiple query.
This validate the effectiveness of learned person parts.

\begin{figure}[tbp]
  \centering
  \includegraphics[width=0.35\paperwidth]{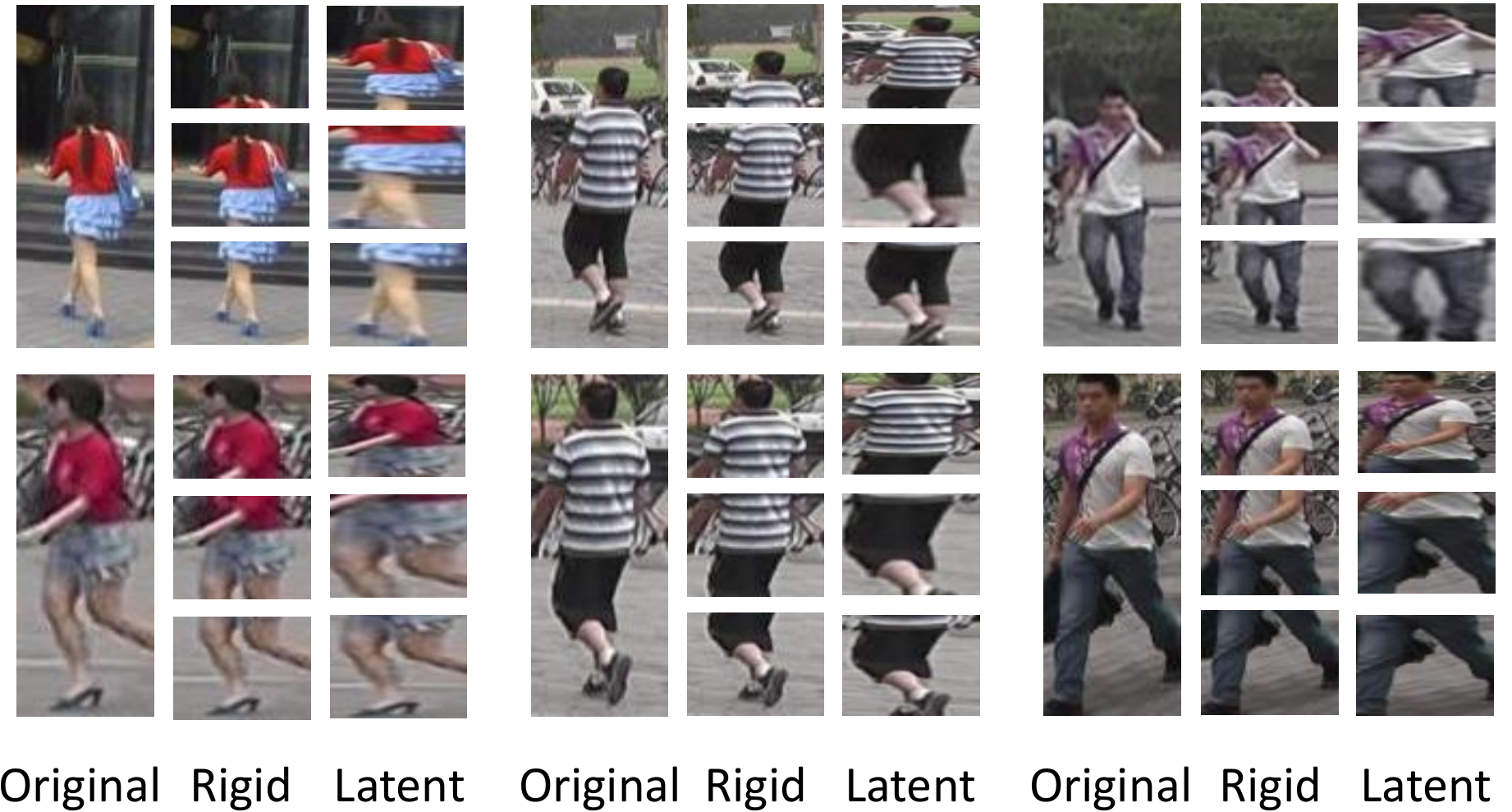}
  \caption{Six samples of original image, rigid predefined parts and learned latent pedestrian parts. Samples in each column are the same person with different backgrounds. Best viewed in color.
  }
  \label{fig:visualizationloc}
  \vspace{-1em}
\end{figure}

For better understanding the learned pedestrian parts, we visualize the localized latent parts in Figure~\ref{fig:visualizationloc} using our fusion model.
For these detected person with large background (the first row in Figure~\ref{fig:visualizationloc}), the proposed model can learn foreground information with complementary latent pedestrian parts.
As we can see, the learned parts consist of three main components, including upper body, middle body (combination of upper body and lower body), and lower body.
Similar results can be achieved when original detection pedestrians contain less background or occlusion (the second row in Figure~\ref{fig:visualizationloc}).
It is easy to see that, the automatically learned pedestrian parts are not strictly head-shoulder, upper body and lower-body.
But it indeed consists of these three parts with large overlap.
Compared with rigid parts, the proposed model can automatically localize the appropriate latent parts for feature learning.

\begin{table}[!htbp]
  \begin{center}
    \scriptsize
    \begin{tabular}{|l|cc|cc|}
    \hline
    Query type & \multicolumn{2}{c|}{Single query} & \multicolumn{2}{c|}{Multiple query} \\
    \hline
    Evaluation metrics & Rank-1    & mAP   & Rank-1    & mAP \\
    \hline
    \hline
    Rigid parts & 72.98  & 49.60  & 82.42  & 60.23  \\
    \hline
    Latent parts & \textbf{76.25}  & \textbf{53.33}  & \textbf{84.12}  & \textbf{62.90}   \\
    \hline
    \end{tabular}%
  \end{center}
  \vspace{-0.5em}
  \caption{Experimental results of rigid parts and learned parts for ReID on the Market1501 dataset.}
  \vspace{-0.0em}
  \label{tab:partmarket}%
\end{table}%

\textbf{Effectiveness of localization loss}
To evaluate the effectiveness of the proposed constraints on the latent part localization network, we conduct additional experiments by adding or deleting proposed $L_{loc}$ in the training stage of body parts network for ReID.
Experimental results are shown in Table~\ref{tab:stnconstrictmarket}.
As we can see, with the additional $L_{loc}$, the Rank-1 accuracy increases by 9.03\%.
We owe the improvements to the effectiveness of the proposed constraints on the part localization network.

\begin{table}[!tp]
  \begin{center}
  \scriptsize
    \begin{tabular}{|l|cc|cc|}
    \hline
    Query type & \multicolumn{2}{c|}{Single query} & \multicolumn{2}{c|}{Multiple query} \\
    \hline
    Evaluation metrics & Rank-1 & mAP   & Rank-1 & mAP \\
    \hline
    \hline
    $L_{cls}$   &   67.22  & 45.27 & 77.55  & 55.40   \\
    \hline
    $L_{cls}+L_{loc}$ & \textbf{76.25}  & \textbf{53.33}  & \textbf{84.12}  & \textbf{62.90}  \\
    \hline
    \end{tabular}%
  \end{center}
  \vspace{-0.5em}
  \caption{The influences of $L_{loc}$ on part-based network on the Market1501 dataset.}
  \vspace{-0.0em}
  \label{tab:stnconstrictmarket}%
\end{table}%

\subsection{Cross-dataset Evaluation}
\label{exp:discussion}

Similar with typical image classification task with CNN, our approach requires large scale of data, not only more identities, but also more instances for each identity.
So we do not train the proposed model on each single small person ReID dataset, such as VIPeR.
Instead, we conduct cross-dataset evaluation from the pretrained model on the Market1501, CUHK03 and MARS datasets to the VIPeR dataset.
The experimental results are shown in Table~\ref{tab:crossviper}.
Compared with other methods, such as Domain Transfer Rank Support Vector Machines~\cite{MaICCV13domain} and DML~\cite{YiICPR14DML},
the models trained on large-scale datasets have better generalization ability and have better Rank-1 identification rate.

To take further analysis of the proposed method, we also fine-tune the model from large dataset Market1501 to small dataset VIPeR.
Experimental results are shown in Table~\ref{tab:markettransfertoviper}.
Our fusion-based model obtains better Rank-1 identification rate than existing deep models, \eg IDLA~\cite{AhmedCVPR15improved} (34.8\%), Gate-SCNN~\cite{VariorECCV16Gated} (37.8\%), SI-CI~\cite{WangfaqiangCVPR16JSC} (35.8\%), and achieves comparable results with DGD~\cite{XiaotongCVPR16Domain} (38.6\%).

\begin{table}[btbp]
  \begin{center}
    \scriptsize
    \begin{tabular}{|l|c|c c c c |}
    \hline
    Methods & Training Set & 1     & 10    & 20    & 30 \\
    \hline
    \hline
    DTRSVM~\cite{MaICCV13domain} & i-LIDS & 8.26  & 31.39  & 44.83  & 53.88  \\
    DTRSVM~\cite{MaICCV13domain} & PRID  & 10.90  & 28.20  & 37.69  & 44.87  \\
    DML~\cite{YiICPR14DML}   & CUHK Campus & 16.17  & 45.82  & 57.56  & 64.24  \\
    \hline
    \hline
    Ours-Fusion  & CUHK03 detected & 17.30  & 44.58  & 55.51  & 61.77  \\
    Ours-Fusion  & CHUK03 labeled &  19.44     &   \textbf{49.99}    &   \textbf{60.78}     &  \textbf{66.74} \\
    Ours-Fusion  & MRAS  & 18.46 &  43.65 &  52.96  &  59.34  \\
    Ours-Fusion  & Market1501 & \textbf{22.21} &  47.24 &  57.13  & 62.26  \\
    \hline
    \end{tabular}%
  \end{center}
  \vspace{-0.5em}
  \caption{Cross-dataset person ReID on the VIPeR dataset}
  \vspace{0em}
  \label{tab:crossviper}%
\end{table}%

\begin{table}[tbp]
  \begin{center}
  \scriptsize
  \begin{tabular}{|l|cccc|}
  \hline
  Method & Rank-1 & Rank-5 & Rank-10 & Rank-20 \\
  \hline
  Our-Part & 32.70  & 57.49 & 67.62 & 78.90 \\
  Our-Body & 33.12 & 60.23 & 72.05 & 82.59 \\
  Our-Fusion & \textbf{38.08} & \textbf{64.14} & \textbf{73.52} & \textbf{82.91} \\
  \hline
  \end{tabular}%
  \end{center}
  \vspace{-0.5em}
  \caption{Experimental results on VIPeR through fine-tuneing the model from Market1501 to VIPeR.}
  \vspace{-1.0em}
  \label{tab:markettransfertoviper}%
\end{table}%

\section{Conclusion}
\label{conclusions}

In this work, we have studied the problem of person ReID in three levels:
1) a multi-scale context-aware network to capture the context knowledge for pedestrian feature learning,
2) three novel constraints on STN for effective latent parts localization and body-part feature representation,
3) the fusion of full-body and body-part identity discriminative features for powerful pedestrian representation.
We have validated the effectiveness of the proposed method on current large-scale person ReID datasets.
Experimental results have demonstrated that the proposed method achieves the state-of-the-art results.

\textbf{Acknowledgement} This work is funded by the National Key Research and Development Program of China (2016YFB1001005), the National Natural Science Foundation of China (Grant No. 61673375, Grant No. 61403383 and Grant No. 61473290), and the Projects of Chinese Academy of Science (Grant No. QYZDB-SSW-JSC006, Grant No. 173211KYSB20160008).

{\small
\bibliographystyle{ieee}
\bibliography{egbib}

\begin{thebibliography}{10}\itemsep=-1pt

\bibitem{AhmedCVPR15improved}
E.~Ahmed, M.~Jones, and T.~K. Marks.
\newblock An improved deep learning architecture for person re-identification.
\newblock In {\em Proc. CVPR}, 2015.

\bibitem{ChendapengCVPR16similarity}
D.~Chen, Z.~Yuan, B.~Chen, and N.~Zheng.
\newblock Similarity learning with spatial constraints for person
  re-identification.
\newblock In {\em Proc. CVPR}, 2016.

\bibitem{Chen2017cvprid}
W.~Chen, X.~Chen, J.~Zhang, and K.~Huang.
\newblock Beyond triplet loss: a deep quadruplet network for person
  re-identification.
\newblock In {\em Proc. CVPR}, 2017.

\bibitem{Chen2017aaai}
W.~Chen, X.~Chen, J.~Zhang, and K.~Huang.
\newblock A multi-task deep network for person re-identification.
\newblock In {\em AAAI}, 2017.

\bibitem{Chengde2016person}
D.~Cheng, Y.~Gong, S.~Zhou, J.~Wang, and N.~Zheng.
\newblock Person re-identification by multi-channel parts-based cnn with
  improved triplet loss function.
\newblock In {\em Proc. CVPR}, 2016.

\bibitem{DingPR15deep}
S.~Ding, L.~Lin, G.~Wang, and H.~Chao.
\newblock Deep feature learning with relative distance comparison for person
  re-identification.
\newblock {\em Pattern Recognition}, 48(10):2993--3003, 2015.

\bibitem{FarenzenaCVPR10}
M.~Farenzena, L.~Bazzani, A.~Perina, V.~Murino, and M.~Cristani.
\newblock Person re-identification by symmetry-driven accumulation of local
  features.
\newblock In {\em Proc. CVPR}, 2010.

\bibitem{FelzenszwalbPAMI10object}
P.~F. Felzenszwalb, R.~B. Girshick, D.~McAllester, and D.~Ramanan.
\newblock Object detection with discriminatively trained part-based models.
\newblock {\em TPAMI}, 32(9):1627--1645, 2010.

\bibitem{Gray07VIPeR}
D.~Gray, S.~Brennan, and H.~Tao.
\newblock Evaluating appearance models for recognition, reacquisition, and
  tracking.
\newblock In {\em IEEE International Workshop on Performance Evaluation for
  Tracking and Surveillance (PETS)}, volume~3, 2007.

\bibitem{GrayECCV08}
D.~Gray and H.~Tao.
\newblock Viewpoint invariant pedestrian recognition with an ensemble of
  localized features.
\newblock In {\em Proc. ECCV}, 2008.

\bibitem{HekaimingCVPR16Residual}
K.~He, X.~Zhang, S.~Ren, and J.~Sun.
\newblock Deep residual learning for image recognition.
\newblock In {\em Proc. ICCV}, 2016.

\bibitem{Ioffe15batch}
S.~Ioffe and C.~Szegedy.
\newblock Batch normalization: Accelerating deep network training by reducing
  internal covariate shift.
\newblock {\em arXiv:1502.03167}, 2015.

\bibitem{JaderbergNIPS15spatial}
M.~Jaderberg, K.~Simonyan, A.~Zisserman, et~al.
\newblock Spatial transformer networks.
\newblock In {\em Proc. NIPS}, 2015.

\bibitem{JiaMM14caffe}
Y.~Jia, E.~Shelhamer, J.~Donahue, S.~Karayev, J.~Long, R.~Girshick,
  S.~Guadarrama, and T.~Darrell.
\newblock Caffe: Convolutional architecture for fast feature embedding.
\newblock In {\em Proc. ACM Multimedia}, 2014.

\bibitem{Jose2016scalable}
C.~Jose and F.~Fleuret.
\newblock Scalable metric learning via weighted approximate rank component
  analysis.
\newblock In {\em Proc. ECCV}, 2016.

\bibitem{KostingerCVPR12}
M.~K{\"o}stinger, M.~Hirzer, P.~Wohlhart, P.~M. Roth, and H.~Bischof.
\newblock Large scale metric learning from equivalence constraints.
\newblock In {\em Proc. CVPR}, 2012.

\bibitem{KrizhevskyNIPS12}
A.~Krizhevsky, I.~Sutskever, and G.~E. Hinton.
\newblock Imagenet classification with deep convolutional neural networks.
\newblock In {\em Proc. NIPS}, 2012.

\bibitem{KviatkovskyPAMI13color}
I.~Kviatkovsky, A.~Adam, and E.~Rivlin.
\newblock Color invariants for person reidentification.
\newblock {\em TPAMI}, 35(7):1622--1634, 2013.

\bibitem{LidangweiArxiv16}
D.~Li, Z.~Zhang, X.~Chen, H.~Ling, and K.~Huang.
\newblock A richly annotated dataset for pedestrian attribute recognition.
\newblock {\em arXiv:1603.07054}, 2016.

\bibitem{LiWeiCVPR14}
W.~Li, R.~Zhao, T.~Xiao, and X.~Wang.
\newblock Deepreid: Deep filter pairing neural network for person
  re-identification.
\newblock In {\em Proc. CVPR}, 2014.

\bibitem{LizhenCVPR13}
Z.~Li, S.~Chang, F.~Liang, T.~S. Huang, L.~Cao, and J.~R. Smith.
\newblock Learning locally-adaptive decision functions for person verification.
\newblock In {\em Proc. CVPR}, 2013.

\bibitem{LiaoshengcaiCVPR15}
S.~Liao, Y.~Hu, X.~Zhu, and S.~Z. Li.
\newblock Person re-identification by local maximal occurrence representation
  and metric learning.
\newblock In {\em Proc. CVPR}, 2015.

\bibitem{LiaoshengcaiICCV15}
S.~Liao and S.~Z. Li.
\newblock Efficient psd constrained asymmetric metric learning for person
  re-identification.
\newblock In {\em Proc. ICCV}, 2015.

\bibitem{LintsungyiECCV14microsoft}
T.-Y. Lin, M.~Maire, S.~Belongie, J.~Hays, P.~Perona, D.~Ramanan,
  P.~Doll{\'a}r, and C.~L. Zitnick.
\newblock Microsoft coco: Common objects in context.
\newblock In {\em Proc. ECCV}, 2014.

\bibitem{Liu2016end}
H.~Liu, J.~Feng, M.~Qi, J.~Jiang, and S.~Yan.
\newblock End-to-end comparative attention networks for person
  re-identification.
\newblock {\em arXiv:1606.04404}, 2016.

\bibitem{MaICCV13domain}
A.~J. Ma, P.~C. Yuen, and J.~Li.
\newblock Domain transfer support vector ranking for person re-identification
  without target camera label information.
\newblock In {\em Proc. ICCV}, 2013.

\bibitem{MatsukawaCVPR16}
T.~Matsukawa, T.~Okabe, E.~Suzuki, and Y.~Sato.
\newblock Hierarchical gaussian descriptor for person re-identification.
\newblock In {\em Proc. CVPR}, 2016.

\bibitem{PaisitkriangkraiCVPR15}
S.~Paisitkriangkrai, C.~Shen, and A.~van~den Hengel.
\newblock Learning to rank in person re-identification with metric ensembles.
\newblock In {\em Proc. CVPR}, 2015.

\bibitem{ProsserBMVC10person}
B.~Prosser, W.-S. Zheng, S.~Gong, T.~Xiang, and Q.~Mary.
\newblock Person re-identification by support vector ranking.
\newblock In {\em Proc. BMVC}, volume~2, page~6, 2010.

\bibitem{SchumannArxiv16deep}
A.~Schumann, S.~Gong, and T.~Schuchert.
\newblock Deep learning prototype domains for person re-identification.
\newblock {\em arXiv:1610.05047}, 2016.

\bibitem{Shihanlin2016Embedding}
H.~Shi, Y.~Yang, X.~Zhu, S.~Liao, Z.~Lei, W.~Zheng, and S.~Z. Li.
\newblock Embedding deep metric for person re-identification: A study against
  large variations.
\newblock In {\em Proc. ECCV}, 2016.

\bibitem{srivastava2014dropout}
N.~Srivastava, G.~E. Hinton, A.~Krizhevsky, I.~Sutskever, and R.~Salakhutdinov.
\newblock Dropout: a simple way to prevent neural networks from overfitting.
\newblock {\em The Journal of Machine Learning Research}, 15(1):1929--1958,
  2014.

\bibitem{SunyiCVPR14deep1}
Y.~Sun, X.~Wang, and X.~Tang.
\newblock Deep learning face representation from predicting 10,000 classes.
\newblock In {\em Proc. CVPR}, 2014.

\bibitem{VariorECCV16Gated}
R.~R. Varior, M.~Haloi, and G.~Wang.
\newblock Gated siamese convolutional neural network architecture for human
  re-identification.
\newblock In {\em Proc. ECCV}, 2016.

\bibitem{Varior2016Siamese}
R.~R. Varior, B.~Shuai, J.~Lu, D.~Xu, and G.~Wang.
\newblock A siamese long short-term memory architecture for human
  re-identification.
\newblock In {\em Proc. ECCV}, 2016.

\bibitem{WangfaqiangCVPR16JSC}
F.~Wang, W.~Zuo, L.~Lin, D.~Zhang, and L.~Zhang.
\newblock Joint learning of single-image and cross-image representations for
  person re-identification.
\newblock In {\em Proc. CVPR}, 2016.

\bibitem{wang2016dari}
G.~Wang, L.~Lin, S.~Ding, Y.~Li, and Q.~Wang.
\newblock Dari: Distance metric and representation integration for person
  verification.
\newblock In {\em AAAI}, 2016.

\bibitem{Wulin2016Personnet}
L.~Wu, C.~Shen, and A.~v.~d. Hengel.
\newblock Personnet: Person re-identification with deep convolutional neural
  networks.
\newblock {\em arXiv:1601.07255}, 2016.

\bibitem{XiaotongCVPR16Domain}
T.~Xiao, H.~Li, W.~Ouyang, and X.~Wang.
\newblock Learning deep feature representations with domain guided dropout for
  person re-identification.
\newblock In {\em Proc. CVPR}, 2016.

\bibitem{XiaotongARXIV16end}
T.~Xiao, S.~Li, B.~Wang, L.~Lin, and X.~Wang.
\newblock End-to-end deep learning for person search.
\newblock {\em arXiv:1604.01850}, 2016.

\bibitem{xu2014person}
Y.~Xu, B.~Ma, R.~Huang, and L.~Lin.
\newblock Person search in a scene by jointly modeling people commonness and
  person uniqueness.
\newblock In {\em Proc. ACM Multimedia}, 2014.

\bibitem{Yang2017aaai}
Y.~Yang, L.~Wen, S.~Lyu, and S.~Z. Li.
\newblock Unsupervised learning of multi-level descriptors for person
  re-identification.
\newblock In {\em AAAI}, 2017.

\bibitem{YangyangECCV14}
Y.~Yang, J.~Yang, J.~Yan, S.~Liao, D.~Yi, and S.~Z. Li.
\newblock Salient color names for person re-identification.
\newblock In {\em Proc. ECCV}, 2014.

\bibitem{YiICPR14DML}
D.~Yi, Z.~Lei, S.~Liao, S.~Z. Li, et~al.
\newblock Deep metric learning for person re-identification.
\newblock In {\em Proc. ICPR}, 2014.

\bibitem{YuKoltun2016}
F.~Yu and V.~Koltun.
\newblock Multi-scale context aggregation by dilated convolutions.
\newblock In {\em Proc. ICLR}, 2016.

\bibitem{ZengArxiv16crafting}
X.~Zeng, W.~Ouyang, J.~Yan, H.~Li, T.~Xiao, K.~Wang, Y.~Liu, Y.~Zhou, B.~Yang,
  Z.~Wang, et~al.
\newblock Crafting gbd-net for object detection.
\newblock {\em arXiv:1610.02579}, 2016.

\bibitem{ZhangLiCVPR16}
L.~Zhang, T.~Xiang, and S.~Gong.
\newblock Learning a discriminative null space for person re-identification.
\newblock In {\em Proc. CVPR}, 2016.

\bibitem{zhang2015bit}
R.~Zhang, L.~Lin, R.~Zhang, W.~Zuo, and L.~Zhang.
\newblock Bit-scalable deep hashing with regularized similarity learning for
  image retrieval and person re-identification.
\newblock {\em TIP}, 24(12):4766--4779, 2015.

\bibitem{ZhangCVPR16sample}
Y.~Zhang, B.~Li, H.~Lu, A.~Irie, and X.~Ruan.
\newblock Sample-specific svm learning for person re-identification.
\newblock In {\em Proc. CVPR}, 2016.

\bibitem{ZhaoruiCVPR13unsupervised}
R.~Zhao, W.~Ouyang, and X.~Wang.
\newblock Unsupervised salience learning for person re-identification.
\newblock In {\em Proc. CVPR}, 2013.

\bibitem{zhaoruiCVPR14learning}
R.~Zhao, W.~Ouyang, and X.~Wang.
\newblock Learning mid-level filters for person re-identification.
\newblock In {\em Proc. CVPR}, 2014.

\bibitem{ZhengliangECCV16}
L.~Zheng, Z.~Bie, Y.~Sun, J.~Wang, C.~Su, S.~Wang, and Q.~Tian.
\newblock Mars: A video benchmark for large-scale person re-identification.
\newblock In {\em Proc. ECCV}, 2016.

\bibitem{ZhengliangICCV15}
L.~Zheng, L.~Shen, L.~Tian, S.~Wang, J.~Wang, and Q.~Tian.
\newblock Scalable person re-identification: A benchmark.
\newblock In {\em Proc. ICCV}, 2015.

\bibitem{zheng2016personreview}
L.~Zheng, Y.~Yang, and A.~G. Hauptmann.
\newblock Person re-identification: Past, present and future.
\newblock {\em arXiv preprint arXiv:1610.02984}, 2016.

\bibitem{ZhengliangArxiv16}
L.~Zheng, H.~Zhang, S.~Sun, M.~Chandraker, and Q.~Tian.
\newblock Person re-identification in the wild.
\newblock {\em arXiv:1604.02531}, 2016.

\bibitem{ZhengPAMI12context}
W.-S. Zheng, S.~Gong, and T.~Xiang.
\newblock Quantifying and transferring contextual information in object
  detection.
\newblock {\em TPAMI}, 34(4):762--777, 2012.

\bibitem{ZhengweishiPAMI13reid}
W.-S. Zheng, S.~Gong, and T.~Xiang.
\newblock Reidentification by relative distance comparison.
\newblock {\em TPAMI}, 35(3):653--668, 2013.

\end{thebibliography}
}

\end{document}